# Counting Coordination Categorially

Crit Cremers and Maarten Hijzelendoorn
Department of General Linguistics
Leiden University
P.O. Box 9515, 2300 RA  Leiden
The Netherlands
e-mail: {cremers, hijzelendrn}@rullet.leidenuniv.nl

## 1. Count invariance and the parsing of coordination

Certain categorial calculi exhibit a property that is known as *count invariance* (Van Benthem 1986). In these grammars, it is true that if a proposition $Y \Rightarrow z$ is derivable, the string to the left of $\Rightarrow$ and the type to its right share the results of a particular way of counting occurrences of basic types. This count protocol discriminates between positive occurences (heads) and negative occurrences (arguments) of basic types; for each basic type $x$ and for each string of types $S$ it yields an integer representing the occurrences of $x$ in $S$.

(1)   Count Protocol

   for each basic type $x$, and for all $y$, $z$ in the set of types closed under left and right division,
   $count_x(x) = 1$
   $count_x(y) = 0$ if $y$ is a basic type and $y \neq x$
   $count_x(y/z) = count_x(y \backslash z) = count_x(y) - count_x(z)$
   $count_x(y_1, ..., y_n) = count_x(y_1) + ... + count_x(y_n)$.

(2)   Count Invariance for $G$

   if $Y \Rightarrow z$ is derivable in some categorial grammar G, then for all basic types $x$, $count_x(Y) = count_x(z)$. By consequence, if $z$ is a basic type, then for all basic types $x \neq z$, $count_x(Y) = 0$.

Count Invariance can be proved for calculi like **L**(ambek) and **A**(djukiewicz/)**B**(ar Hillel), but does, of course, not hold in systems that exhibit Contraction - $y\ y \Rightarrow y$ - and/or Monotonicity - if $X \Rightarrow y$ then $X\ x \Rightarrow y$; cf. Van Benthem (1991). Furthermore, obeying Count Invariance is not sufficient condition for derivability. For example: $(x/y)/z\ y\ z \Rightarrow x$ and $x \backslash y\ y \Rightarrow x$ comply with Count Invariance, but are not derivable in **L** or **AB**.

Parsing grammars based on calculi that exhibit Count Invariance, may gain profit from this property, because it can be used to delimit the search space. By contraposition of (2), a proposition $Y \Rightarrow z$ cannot be proved in a count invariant system G if for some basic type $x$, $count_x(Y) \neq count_x(z)$. Whenever a string of words gives rise to more than one sequence of types the reducibility of which is checked, one may try to cancel some of these sequences by testing them against Count Invariance; see e.g. Moortgat (1988) and Cremers (1989). The number of possible sequences of types for a sentence equals the cardinality of the Cartesian product over the sets of lexical categories of the words, and is exponentially dependent on the length of the sentence. Since only relatively few assignments represent viable options, efficiency requires serious pruning.

Almost by definition, coordination in natural language involves the multiplication of types: a certain subsequence of types to the left of the coordination point is doubled or mirrored at the right of the coordination point. In some languages, like Dutch, there are hardly any limitations as to the nature of the repeated subsequence (Houtman 1994, but see Grootveld 1994 for a grammaticalization). The relative unrestrictedness of coordination is reflected in the proposal (e.g. Moortgat 1988, Wittenburgh 1986, Steedman 1990) to categorize coordinating elements like *and* as $(x\backslash x)/x$, i.e. by means of essential variables over types.

At the same time, this categorization accounts for the doubling of types induced by coordination. This doubling would interfere with count invariance: the count of certain types will be unbalanced by the repetition of subsequences. The variables in the coordination type are supposed to be instantiated with a type that can be assigned to the subsequence that is coordinated. Since the type for *and* divides the count for *x* by 2, the effect of doubling is neutralized and count invariance is still a property of a coordinated sentence. Therefore, the following statement is valid.

(3) If $Y\ (x\backslash x)/x\ Z \Rightarrow s$ is derivable, then
 (a) there is a partition $Y'\ C1$ of $Y$ and a partition $C2\ Z'$ of $Z$, and
 (b) for some $c$, both $C1 \Rightarrow c$ and $C2 \Rightarrow c$ are derivable, and
 (c) $Y'\ c\ (c\backslash c)/c\ c\ Z' \Rightarrow s$ obeys count invariance.

This strategy for dealing with coordination, however, makes Count Invariance a blunt knife in pre-parsing. For a given sequence $Y\ x\backslash x/x\ Z$ we can check count invariance only by guessing the partition $Y'\ C1$ of $Y$ and $C2\ Z'$ of $Z$ independently of each other and by guessing a type $c$ such that both $C1$ and $C2$ reduce to it. The space for each guess is finite (Van Benthem 1991; p. 77), but the indeterminacy of the triple guess disqualifies this approach for an effective application of Count Invariance: trying to find out what should be counted and tested for invariance presupposes partial parsing of the sequence in a nondeterministic mode. The partition of the left and right substrings and the type of the coordinated strings are supposed to be part of the outcome of the structural parsing, and not to be part of the preprocessing.

In the presence of coordination, Count Invariance (2) is of no help at selecting viable sequences prior to parsing. Therefore, we developed a weaker alternative which can be effectively exploited in delimiting the search space for coordinated sentences.

## 2. An operational count invariant for coordinated sequences

Coordination can be constructed so as to imply that certain elements outside the scope of coordination have a double task with respect to elements inside the scope of coordination: they have to serve elements in both coordinates. For example, in a string $x/y\ y\ \&\ y\ z\backslash x \Rightarrow z$ the negative occurrence of $y$ in $x/y$ has to account for the two positive occurrences of $y$ to the left and the right of the coordinator $\&$. In fact, $y$ is coordinated in that string. In the same vein, the positive occurrence of $u$ in $x/y\ y/u\ \&\ y/u\ u\ z\backslash x \Rightarrow z$ has to compensate for the two negative occurrences of $u$ that are within the scope of coordination. If such a positive $u$ were not available or were occupied, as in $x/y\ y/u\ \&\ y/u\ u\ z\backslash x\backslash u \Rightarrow z$, coordination is bound to fail. Consequently, by just counting the positive and negative occurrences of primitive types to the left and the right of the coordinator, we can check whether double function categories are available. All we have to know are the count values of the intended coordinated substring, since these values will specify the nature of the need for double function primitives. The question, then, is to determine these 'coordinated' count values without parsing.

Suppose we have a string of the form $W_1$ *and* $W_2$. Let L and R be assignments of types to $W_1$ and $W_2$, respectively, that have not been ruled out by other occurrence checks, which are independent of coordination and mainly rely on the directionality invariant (see e.g. Steedman 1990 and Cremers 1989). Let furthermore with L be associated a register $reg_L$ of quadruples of integers <*sathead, satarg, freehead, freearg*>, such that for each primitive type $x$ there is an $x$-quadruple specifying:

> $sathead^x_L$: the number of positive occurrences of $x$ in L that are saturated by negative occurrences of $x$ under a \-slash in L;
> $satarg^x_L$: the number of negative occurrences of $x$ under a /-slash in L that are saturated by positive occurrences of $x$ in L;
> $freehead^x_L$: the number of positive occurrences of $x$ in L that are not saturated by negative occurrences of $x$ in L;
> $freearg^x_L$: the number of negative occurrences of $x$ under a /-slash in L that are not saturated by positive occurrences of $x$ in L.

A head (or positive occurrence of) $x$ is counted as free in L if there is no /$x$ to its left or \$x$ to its right in L by which it could be saturated. Similarly, an argument (or negative occurrence of) /$x$ is free in L if no head to its right can possibly saturate it. Arguments \$x$ cannot be free in L, by the directionality invariant. For R, a similar register $reg_R$ is supposed to be available, though directional parameters are reversed. What do the registers tell us about the possibility of combining L and R into one hypothesis $L \& R \Rightarrow s$?

The values for $freehead^x$ and $freearg^x$ in each quadruple provide the number of positive and negative occurrences, respectively, that are not matched at their side of the coordinator. These occurrences may or may not be in the domain of coordination. Suppose an unsaturated occurrence of $x$ in L is in the scope of coordination. This occurrence, then, has a parallel occurrence in R that is necessarily saturated at its side (Cremers 1993; ch. 3). For an unsaturated occurrence of $x$ in L to be inside the scope of coordination, a potentially matching type must be available in the saturated part of R. By the same line of reasoning, if the occurrence of $x$ that is unsaturated in L, is to be outside the scope of coordination, it has to match some unsaturated occurrence in R. We can illustrate the range of possibilities with a simple example.

> (4)  $X\ x\ Y\ \&\ W\ x\ V\ y\backslash x\ Z \Rightarrow s$
>  L = $X\ x\ Y$
>  R = $W\ x\ V\ y\backslash x\ Z$
>  $x$-quadruple in $reg_L$ = <0, 0, 1, 0>
>  $x$-quadruple in $reg_R$ = <0, 1, 0, 0>

The $x$-quadruple in $reg_L$ tells us that some positive occurrence of $x$ is free in L: $freehead^x_L$ = 1. This occurrence may be part of the coordinated substring. If it is, there must be some negative occurrence in R that takes care of $x$ in L and of its counterpart in R. This negative occurrence, however, is saturated in R, and must be accounted for in the number of saturated arguments \$x$ in $reg_R$, $satarg^x_R$; this number happens to be 1, due to the

composition of R . Now suppose the positive unsaturated occurrence of $x$ in L is not inside the scope of coordination. Then there must be a negative occurrence $\backslash x$ in R that is not yet saturated. This negative occurrence should be accounted for in the number *freearg*$^x_R$. Since, in the example (4), this number is 0, all the occurrences counted in *freehead*$^x_L$ must be in the scope of coordination.

This type of reasoning can be generalized in order to decide, under the hypothesis that the proposition is derivable, how many unsaturated negative and positive occurrences in L and R must be in the scope of coordination.

Let $\lambda_x$ be the difference *freehead*$^x_L$ - *freearg*$^x_R$, for some basic type $x$. Clearly, if $\lambda_x > 0$, there are $\lambda_x$ positive occurrences of $x$ in L that are not matched by negative occurrences $\backslash x$ free in R. These $\lambda_x$ occurrences must therefore be in the scope of coordination and be co-covered by already saturated negative occurrences $\backslash x$ in R. In that case, *satarg*$^x_R$ must be at least as large as $\lambda_x$. On the other hand, if $\lambda_x < 0$, there must be negative occurrences $\backslash x$ in R that, for the string to be grammatical, must be matched by already saturated occurrences $x$ in L, accounted for in the number *sathead*$^x_L$; this number must be large enough to accommodate the $|\lambda_x|$ negative occurrences $\backslash x$ in R. If $\lambda_x = 0$, all or none of the *freehead*$^x_L$ positive occurrences and *freearg*$^x_R$ negative occurrences are in the scope of coordination, depending on other parameters. An equivalent reasoning can be built around the value $\rho_x$, being the difference *freehead*$^x_R$ - *freearg*$^x_L$. (As a corollary, $\lambda_x + \rho_x = count_x(C)$ for that proper affix C of L and of R that happens to be in coordination.)
Thus we have the following inequalities for two assignments L and R:

(5) Coordinative Count Invariant

if $L\ \&\ R \Rightarrow s$ is derivable,
then for every primitive type $x \neq s$ such that
⟨*sathead*$^x_L$, *satarg*$^x_L$, *freehead*$^x_L$, *freearg*$^x_L$⟩ is in $reg_L$ and
⟨*sathead*$^x_R$, *satarg*$^x_R$, *freehead*$^x_R$, *freearg*$^x_R$⟩ is in $reg_R$ and
$\lambda_x$ = *freehead*$^x_L$ - *freearg*$^x_R$ and
$\rho_x$ = *freehead*$^x_R$ - *freearg*$^x_L$, it is true that
$\lambda_x \leq$ *satarg*$^x_R$ and $\rho_x \leq$ *satarg*$^x_L$ and
$-\lambda_x \leq$ *sathead*$^x_L$ and $-\rho_x \leq$ *sathead*$^x_R$.

By contraposition, a string with a coordinator is not reducible to *s* if for some primitive type $x$ one or more of the inequalities does not hold. This justifies the following statement:

(6) Conjoinability

An assignment L of types to the words to the left of a coordinator and an assignment R of types to the words to its right are *conjoinable with respect to a basic type x* iff the quadruples for $x$ in $reg_L$ and $reg_R$ satisfy the inequalities of (5).
Two strings of types L and R are *conjoinable* as $L\ \&\ R$ iff L and R are conjoinable with respect to every basic type $x$, $x \neq s$.

Given a set of assignments LL and a set of assignments RR, with registers associated to each member of each set, it is easy to select those pairs ⟨L, R⟩ in LL × RR that are conjoinable. Only these pairs are transmitted to the proper parsing module.

The selection of conjoinable pairs of strings is completely deterministic. While an assignment is built, a register is kept and associated with it. If the assignment passes some occurrence tests that are not at stake here, it is admitted to LL or RR; its register is fixed. The procedure for comparing assignments in the product LL × RR checks pairs of quadruples from the registers in the spirit of the contraposition to Coordinative Count Invariant (5). This requires only a fixed number of steps per pair of quadruples and per pair of registers.

### 3. The effectivity of Conjoinability

The conjoinability test (6) has been implemented in a categorial parsing system called Delilah, which embodies among other things the algorithm for parsing unbounded coordination of Cremers (1993). Here is an example of the effect of conjoinability in this system. Sentence (7) contains 51 words from a restricted but highly polymorphic lexicon of Dutch. In this lexicon, all kinds of combinatorial options for each word are spelled out as categorial types. Because of this lexical polymorphism, the number of possible different assignments of lexical categories to the sentence is 4.6e12. The coordinator *en* is handled syncategorimatically.

> (7)  Omdat ik niet had willen zeggen dat ik door de man met
> *Because I not had want say that I by the man with*
> de auto werd gedwongen te proberen Henk met de pop **en**
> *the car was forced to try Henk with the doll and*
> Agnes met een boek te laten spelen, werd de man door de
> *Agnes with a book to let play, was the man by the*
> vrouw gedwongen te zeggen dat hij mij niet met de pop
> *woman forced to tell that he me not with the doll*
> wilde proberen te laten spelen.
> *wanted try to let play*.

First, some general occurrence checks that are not related to coordination (cf. Cremers 1989) keep the number of 4.6e12 possibilities just virtual by reducing dynamically the set of viable assignments of categories to this string to 1.1e5, or 2.6e-6 % of the number of possibilities. This number is the product of 736 assignments to the left - the set LL - and 160 assignments to the right of the coordinator - the set RR. Applying Conjoinability to LL × RR, leaves 1256 combinations of a left and a right assignment as viable, i.e. 1.1 % of |LL × RR|. Only these 1256 assignments are analyzed by the parser, which in this case finds one or more derivations for three of them.
By the joined forces of independent occurrence checks and Conjoinability the number of assignments admitted to full parsing, is thus reduced to 2.8e-8 % of the original 4.6e12.

Since it is very hard, if not impossible, to find a general metric for the effect of Conjoinability (6) on the set of admitted sequences, we give just some more data. Figure (8) holds, for some grammatical sentences over the same lexicon as sentence (7), their length L, the number of possible assignments PA, the product CP = |LL × RR|, the ratio CP/PA as a percentage, the number AA of assignments admitted to parsing, and the ratio AA/CP as a percentage. The latter ratio measures the effectivity of applying Conjoinability. The ratio AA/PA gives the percentage of the total space of possibilities that is transmitted to the parsing module; it stands for the effectiveness of the count procedure as a whole.

(8)

| L  | PA    | CP    | CP/PA% | AA | AA/CP% | AA/PA% |
|----|-------|-------|--------|----|--------|--------|
| 16 | 6.0e3 | 2e1   | 3.3e-1 | 2  | 1.0e1  | 3.3e-2 |
| 22 | 7.7e5 | 9.6e1 | 1.2e-2 | 4  | 4.2e0  | 5.2e-4 |
| 33 | 5.0e7 | 1.2e3 | 2.3e-3 | 8  | 6.9e-1 | 1.6e-5 |
| 39 | 7.4e7 | 5.4e3 | 7.3e-3 | 72 | 1.3e0  | 9.7e-5 |
| 44 | 4.4e8 | 2.2e4 | 5.1e-3 | 2  | 8.9e-3 | 4.5e-7 |

A comparable overview is given in (9) for some ungrammatical sentences. The zero values mean that the system, in a stage prior to proper parsing, could not find any potentially derivable sequence; in the case of ungrammatical sentences this is, of course, a desirable result.

(9)

| L  | PA    | CP    | CP/PA% | AA  | AA/CP% | AA/PA% |
|----|-------|-------|--------|-----|--------|--------|
| 15 | 2.0e3 | 2.4e1 | 1.2e0  | 0   | 0      | 0      |
| 22 | 9.6e2 | 2.2e2 | 2.3e1  | 4   | 1.9e0  | 4.2e-1 |
| 33 | 1.5e6 | 2.1e3 | 1.4e-1 | 0   | 0      | 0      |
| 38 | 3.7e7 | 2.7e3 | 7.3e-3 | 24  | 8.8e-1 | 6.5e-5 |
| 47 | 2.6e8 | 6.7e4 | 2.6e-2 | 242 | 3.6e-1 | 9.2e-5 |

It is our impression that the percentage AA/PA, measuring the effectiveness of the complete battery of count checks prior to parsing, including Conjoinability, tends to decrease as PA, the number of lexically possible assignments, increases.

In many cases, Conjoinability admits more assignments to the parser module than is necessary or desirable from a parsing point of view. The remaining redundancy is due to the fact that the Coordinative Count Invariant (5) is stated in terms of inequalities, rather than in terms of equalities. Count invariants in terms of inequalities, however, are the best we can get for a pruning instrument prior to parsing in the presence of coordination. To see why, consider the family of conjunctions

(10) Henk zei dat ik Agnes een boek **en**
*Henk said that I Agnes a book and*
(a) een tijdschrift had gegeven
   *a magazine had given*
(b) de vrouw een tijdschrift had gegeven
   *the woman a magazine had given*
(c) jij de vrouw een tijdschrift had gegeven
   *you the woman a magazine had given*

Each of the right hand sides (10)(a) - (c) are legitimate continuations of the given left hand side. Now take some assignment L to the left environment of the coordinator, i.e. to *Henk zei dat ik Agnes een boek*. From L alone one cannot make any predictions as to the nature of assignments R that are conjoinable. The only condition imposed by L on R is that some proper prefix of R should reflect some proper suffix of L. Because coordination does not express itself functionally to the left or to the right of the coordinator, we cannot tell which suffixes of L are available. In general, then, there will be more than one conceivable way of grammatically extending the string to the right of the coordinator. Consequently, several essentially different sequences R for (10)(a) - (c) have to be conjoinable to L. Nothing in L or its register imposes at forehand occurrence conditions on assignments to possible extensions of the string to the right of a coordinator. The fact is that an assignment L with a fixed register $reg_L$ may be conjoinable with many different Rs. Since these Rs have all different registers, it is impossible to derive an nontrivial zero-condition on $reg_R$ from $reg_L$ - nor can it be done the other way around. One would have to look for a two-place function *f* such that *f(l,r)* is constant while *r* varies: these functions will hardly be dependent on *r* in an interesting way.

The Coordinative Count Invariant defines the margins for the candidate Rs as narrowly as possible, but has to leave room for variation caused by the functional indeterminacy of coordination.